\pdfoutput=1
\documentclass[11pt]{article}
\usepackage[preprint]{acl}

\usepackage{times}
\usepackage{latexsym}
\usepackage{amsmath}
\usepackage{multirow}
\usepackage{amsfonts}
\usepackage[T1]{fontenc}

\usepackage[utf8]{inputenc}
\usepackage{subcaption}

\usepackage{microtype}

\usepackage{inconsolata}
\usepackage[most]{tcolorbox}
\usepackage{xcolor}
\usepackage{footmisc}
\definecolor{errorcolor}{RGB}{192, 0, 0} 

\newtcolorbox{casebox}[2]{
    enhanced,
    breakable,
    colback=black!3!white,
    colframe=black!60!white, 
    fonttitle=\bfseries,
    title=Case Study -- #1: \textcolor{errorcolor}{#2},
    attach boxed title to top left={yshift=-2mm, xshift=3mm},
    boxed title style={
        colback=black!80,
        arc=2mm,
        outer arc=2mm
    },
    segmentation style={solid, black!50!white, line width=1pt},
    drop shadow={black!50!white}
}
\usepackage{graphicx}
\usepackage{booktabs}
\usepackage{enumitem}

\usepackage{xcolor}
\definecolor{skyblue}{RGB}{0,0,255}   
\definecolor{orange}{RGB}{255,165,0} 
\definecolor{green}{RGB}{0,205,0}
\definecolor{ForestGreen}{RGB}{34,139,34}

\newcommand{\MathOlympiadEval}{\textsc{MathOlympiadEval}}

\newcommand\blfootnote[1]{%
  \begingroup
  \renewcommand\thefootnote{}\footnote{#1}%
  \addtocounter{footnote}{-1}%
  \endgroup
}

\title{Right Is Not Enough: The Pitfalls of Outcome Supervision in Training LLMs for Math Reasoning}

\author{
Jiaxing Guo\textsuperscript{1,2}, \quad
Wenjie Yang\textsuperscript{1,2}, \quad
Shengzhong Zhang\textsuperscript{3}, \quad
Tongshan Xu\textsuperscript{1} \\
Lun Du\textsuperscript{2}\thanks{Corresponding author.}, \quad
Da Zheng\textsuperscript{2}\footnotemark[1], \quad
Zengfeng Huang\textsuperscript{1}\footnotemark[1] \\
\\
\textsuperscript{1} School of Data Science, Fudan University \qquad \textsuperscript{2} Ant Group \\
\textsuperscript{3} College of Computer Science and Technology, Nanjing University of Aeronautics and Astronautics \\
\\
\texttt{jxguo24[at]m.fudan.edu.cn}
}

\begin{document}
\maketitle

\blfootnote{Preprint. Under review. Code and dataset will be made public soon.}

\begin{abstract}
Outcome-rewarded Large Language Models (LLMs) have demonstrated remarkable success in mathematical problem-solving. However, this success often masks a critical issue: models frequently achieve correct answers through fundamentally unsound reasoning processes, a phenomenon indicative of reward hacking. We introduce \textbf{MathOlympiadEval}, a new dataset with fine-grained annotations, which reveals a significant gap between LLMs' answer correctness and their low process correctness. Existing automated methods like LLM-as-a-judge struggle to reliably detect these reasoning flaws. To address this, we propose \textbf{ParaStepVerifier}, a novel methodology for meticulous, step-by-step verification of mathematical solutions. ParaStepVerifier identifies incorrect reasoning steps. Empirical results demonstrate that ParaStepVerifier substantially improves the accuracy of identifying flawed solutions compared to baselines, especially for complex, multi-step problems. This offers a more robust path towards evaluating and training LLMs with genuine mathematical reasoning.
\end{abstract}

\section{Introduction}

\begin{figure}[t]
    \centering 
        \centering 
        \includegraphics[width=\linewidth]{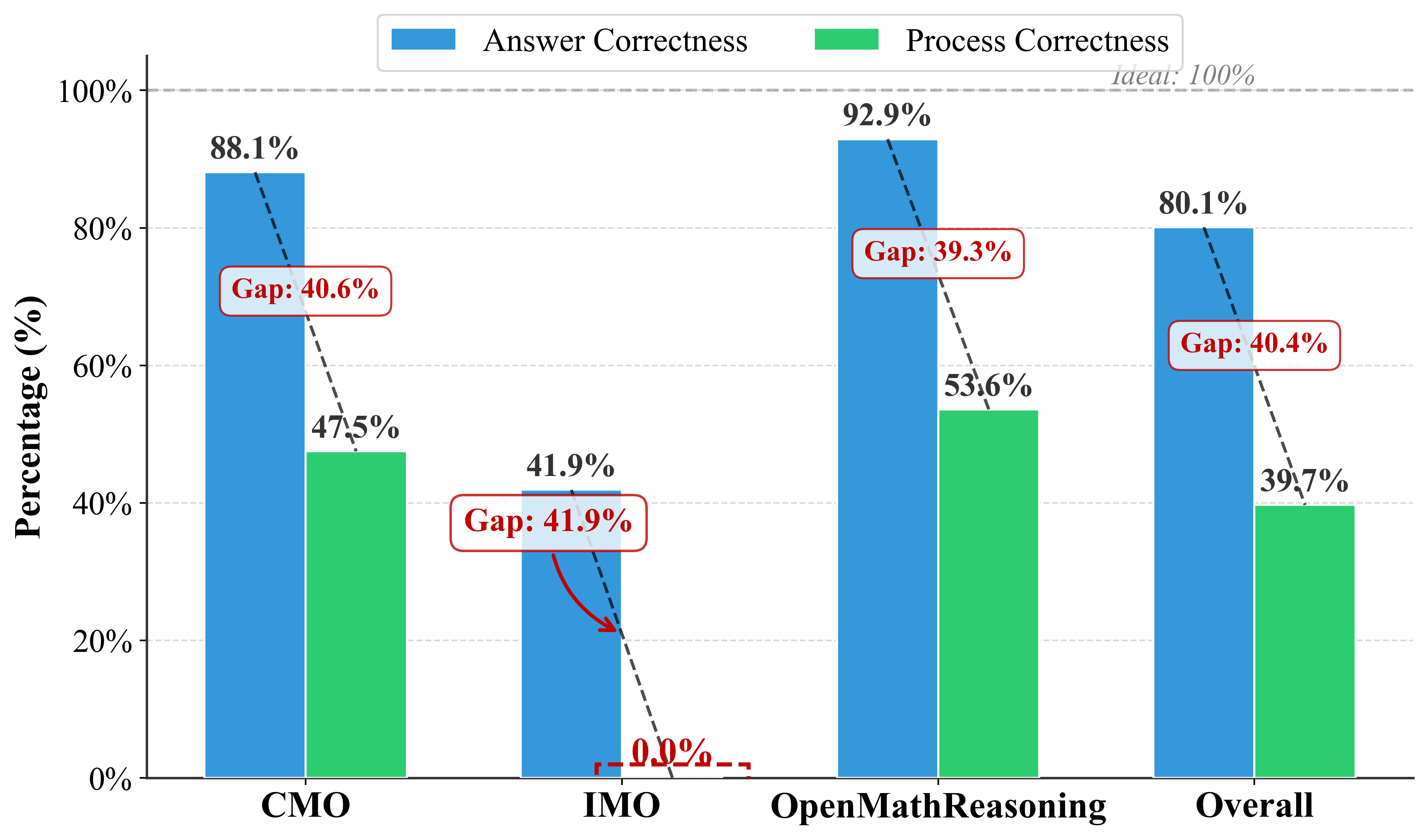}
    \caption{Large Language Models frequently achieve correct mathematical answers through fundamentally unsound reasoning processes. Human evaluation on \textsc{MathOlympiadEval} reveals a significant disparity between high answer correctness and low process correctness.}
    \label{fig:answer_process_gap}
\end{figure}

Complex mathematical reasoning represents a critical frontier in artificial intelligence, where rigorous evaluation methodologies are essential for achieving meaningful progress. Large Language Models (LLMs) are increasingly employed for mathematical problem solving, typically trained through Supervised Fine-Tuning (SFT) and Reinforcement Learning (RL) optimized for final outcomes \cite{liu2024deepseek,guo2025deepseek}. Models such as DeepSeek-R1 \cite{guo2025deepseek} and QwQ-32B \cite{team2025qwq}, which rely on outcome-based reward optimization, have achieved impressive performance on mathematical benchmarks. These successes have led some to suggest that outcome-focused optimization alone may suffice, potentially settling the long-standing debate between outcome-based and process-based reward strategies.

However, our meticulous human evaluation of LLM outputs on \textsc{MathOlympiadEval}—a diverse dataset we construct comprising challenging mathematical problems with LLM-generated solutions annotated for both final answer correctness and fine-grained reasoning validity—reveals a critical disconnection: models frequently arrive at correct final answers through fundamentally flawed reasoning processes. Specifically, while reasoning-oriented models such as DeepSeek-R1 and QwQ-32B attain \textbf{80.1\%} answer accuracy, human experts judge only \textbf{39.7\%} of the corresponding reasoning paths to be sound.  In other words, among the problems considered ``correct'' based on final answers, \textbf{fewer than half} are truly correct when scrutinized for reasoning validity (Figure~\ref{fig:answer_process_gap}). This phenomenon is not confined to a narrow subset of problems, but is widespread across tasks drawn from major mathematical competitions, such as the International Mathematical Olympiad (IMO), the Chinese Mathematical Olympiad (CMO), and community-driven platforms like Art of Problem Solving (AoPS). Across these diverse sources, LLMs consistently exhibit a pattern of producing seemingly correct answers underpinned by logically unsound reasoning.

Such findings strongly suggest that optimizing solely for final outcomes can incentivize models to adopt "shortcut" strategies—effectively engaging in reward hacking—learning to maximize outcome metrics without developing genuine reasoning integrity. This issue is further compounded by the risk of systemic propagation. Flawed solution processes symptomatic of reward hacking may appear in datasets used for supervised fine-tuning (SFT), potentially perpetuating undesirable reasoning patterns in subsequent model generations. Our analysis of samples from OpenMathReasoning~\cite{moshkov2025aimo2winningsolutionbuilding}, which are reportedly used in SFT pipelines, reveals that \textbf{57.7\%} of the solution traces contain reasoning flaws. This raises serious concerns about a feedback loop in which erroneous model outputs contaminate training data, reinforcing shallow or incorrect reasoning and ultimately limiting the models' capacity for reliable mathematical problem solving.

These results indicate that existing benchmarks inadequately assess the reasoning abilities of Large Language Models (LLMs), as outcome-only metrics often fail to capture critical process-level flaws. Scalable LLM-as-a-Judge approaches—while increasingly adopted for automatic evaluation—face substantial reliability challenges~\cite{faggioli2023reliability, shi2024reliablejudge}. These automated judges suffer from a range of issues—including prompt sensitivity and inconsistency~\cite{sclar2023prompt, yin2024promptagnostic}, and stylistic bias and self-preference~\cite{zheng2023judging, panickssery2024selfpreference}—ultimately showing weak correlation with human judgments~\cite{zheng2023judging, wang2024alingmentstability}. 

Therefore, despite the apparent successes of outcome-rewarded models, there remains a pressing need for evaluators that assess the reasoning process itself. Moving beyond mere answer checking is essential to detect and penalize reward hacking effectively. To bridge this gap, we introduce ParaStepVerifier—an agent designed for fine-grained, step-by-step verification of mathematical solutions. ParaStepVerifier systematically identifies flawed reasoning, pinpointing the exact steps where logic breaks down. This granular feedback not only enables detailed diagnostics of failure modes, but also supports the curation of higher-quality training data. By providing such structured insights, ParaStepVerifier offers a more reliable foundation for both model evaluation and the development of more robust mathematical reasoning in LLMs.

Our main contributions are:
\begin{itemize}[itemsep=0pt, topsep=0pt]
    \item We construct \textbf{MathOlympiadEval}, a diverse dataset of challenging mathematical problems sourced from CMO, IMO, and OpenMathReasoning. Each problem is annotated by human experts for reasoning correctness.
    \item We uncover a significant gap between answer accuracy (80.1\%) and reasoning correctness (39.7\%) in outcome-rewarded LLMs (e.g., DeepSeek-R1) when performing mathematical reasoning, indicating that correct answers often arise from flawed reasoning. We further identify common failure patterns—such as guessing and logical errors—and show that 57.7\% of sampled SFT data (e.g., OpenMathReasoning) contain such flaws, raising concerns about reward hacking and reasoning degradation through data contamination.
    \item We propose \textbf{ParaStepVerifier}, an agent performing step-by-step verification and detailed error identification. It can evaluate mathematical reasoning processes, detect reward hacking, and is applicable to both answer-finding and proof-based tasks.
\end{itemize}

\section{Related Work}

\paragraph{LLM-as-a-judge} Existing benchmarks inadequately evaluate the reasoning processes of Large Language Models (LLMs), as outcome-only metrics often overlook critical process-level flaws.
Scalable LLM-as-a-Judge approaches, while increasingly used, face significant reliability challenges~\cite{faggioli2023reliability, shi2024reliablejudge}.
These automated judges exhibit persistent issues, including: inconsistencies and prompt sensitivity~\cite{sclar2023prompt, yin2024promptagnostic}; inherent biases toward verbosity or specific styles, and nepotism favoring outputs stylistically similar to their own~\cite{zheng2023judging, panickssery2024selfpreference}; a tendency to overlook logical fallacies~\cite{tonmoy2024factualitysurvey, zheng2023judging}; poor correlation with human judgments~\cite{zheng2023judging, wang2024alingmentstability}; and an inability to detect errors when the judge shares weaknesses with the assessed model~\cite{tonmoy2024factualitysurvey, openai2023gpt4}.

\paragraph{Reasoning Models}
Critical gaps exist in evaluating LLM mathematical reasoning. Aligning with \citet{wu2025proof}'s findings on LLM proof struggles and unreliable LLM-as-a-judge evaluations, our \textsc{MathOlympiadEval} dataset broadly documents the answer-process correctness discrepancy. To address this, \textbf{ParaStepVerifier} offers robust step-by-step verification. Similarly, \citet{li2025brains} identified reward hacking in Olympiad problem-solving; \textsc{MathOlympiadEval} extends this finding, demonstrating its pervasiveness across diverse mathematical competitions, including high-school levels (Table~\ref{tab:reasoning_analysis_combined_146}).

The reliance on outcome-based rewards for training verifiers, such as Heimdall \cite{liu2025heimdall}, is also challenged by our research, as \textsc{MathOlympiadEval} shows this supervision masks underlying reasoning flaws. While StepMathAgent by \citet{wang2025stepmathagent} proposes a step-wise scoring system, our \textbf{ParaStepVerifier} focuses on definitively establishing the binary correctness of each reasoning step.

\section{\textsc{MathOlympiadEval}: A Dataset for Mathematical Reasoning and Its Evaluation}
\label{sec:dataset_and_eval}

This section introduces \textsc{MathOlympiadEval}, a new dataset developed to facilitate a nuanced evaluation of mathematical reasoning in Large Language Models (LLMs). We first define the core reasoning tasks the dataset addresses, then detail its construction and rigorous annotation process. Finally, we present initial evaluation results obtained using \textsc{MathOlympiadEval}, highlighting critical discrepancies between models' answer correctness and the soundness of their reasoning.

\subsection{Task Definition}
\label{sec:task_definition_dataset}

Given a mathematical problem $P$ from the \textsc{MathOlympiadEval} dataset and a candidate solution $S$ generated by an LLM, where $S$ consists of a sequence of reasoning steps $\{s_1, s_2, \ldots, s_n\}$ culminating in a final answer $A$, the evaluation task aims to assess two distinct aspects of $S$:
\begin{enumerate}
\item \textbf{Answer Correctness} ($C_A \in \{0, 1\}$): For answer-finding problems, we determine whether the final answer $A$ matches the reference answer $R$. This binary judgment represents the traditional outcome-based evaluation commonly used in mathematical benchmarks.
\item \textbf{Reasoning Correctness} ($C_R \in \{0, 1\}$): We assess whether the solution process follows valid mathematical principles without logical gaps, incorrect operations, or unjustified conclusions. A solution has correct reasoning if and only if each step $s_i$ logically follows from the problem statement and preceding steps $(P, \{s_1, \ldots, s_{i-1}\})$.
\end{enumerate}

\subsection{\textsc{MathOlympiadEval} Construction and Annotation}
\label{sec:dataset_construction_annotation_detailed}

\paragraph{Problem Curation}
The \textsc{MathOlympiadEval} dataset comprises 204 high-quality mathematical problems drawn from four main categories: the China Mathematical Olympiad Preliminary Contest (CMO-Preliminary; 86 problems), the International Mathematical Olympiad (IMO; 62 problems), and NVIDIA's OpenMathReasoning dataset (56 problems) ~\cite{moshkov2025aimo2winningsolutionbuilding}. To ensure problem diversity and mitigate selection bias, problems from CMO-Preliminary, IMO, and OpenMathReasoning (specifically from the Art of Problem Solving "High School Olympiads" section) were randomly sampled. The curated dataset includes 58 proof-based problems and 146 answer-finding problems, representing a range of typical challenges in competition-level mathematics.

\paragraph{LLM Solution Generation for the Dataset}
Candidate solutions for the problems in \textsc{MathOlympiadEval} were generated using two advanced open-source LLMs: QwQ-32B and DeepSeek-R1, both fine-tuned using Reinforcement Learning with outcome-based rewards. A consistent prompting methodology ensured that model outputs included complete reasoning steps. For proof-based problems, models generated formal proofs and reasoning processes. For answer-finding problems, models provided both the final answer and a detailed solution path. These generated solutions form part of the dataset alongside the problems.

\paragraph{Annotation Methodology}
The LLM-generated solutions within \textsc{MathOlympiadEval} underwent a rigorous annotation process. First, an initial automated verification was performed for answer correctness where applicable. Subsequently, a human annotation phase was conducted by mathematics graduate students with extensive experience in mathematical competitions. Annotators were provided with the preliminary verification results and official reference solutions. All annotators received uniform training and guidance.

To gain granular insights into model deficiencies, solutions identified as incorrect in their reasoning were systematically categorized. This error categorization is a key feature of the dataset's annotations. The predefined error types include:
\begin{itemize}
    \item \textbf{Solution by Guess:} Detecting instances where solutions appear to rely on specific examples, pattern-spotting without rigorous generalization, or other non-deductive leaps to reach a conclusion.
    \item \textbf{Circular Reasoning:} Identifying arguments where the proposition to be proven is implicitly or explicitly assumed within the premises.
    \item \textbf{Inequality Manipulation Errors:} Detecting incorrect applications or transformations of inequality properties and operations.
    \item \textbf{Calculation Errors:} Identifying arithmetic or algebraic mistakes within the solution steps.
    \item \textbf{Logical Fallacies:} Detecting broader flawed reasoning patterns or formal/informal fallacies not specifically covered by other categories.
\end{itemize}

Each annotation, including error categorization, underwent a three-round iterative review process, with each round conducted by a different expert annotator to ensure the reliability and accuracy of the labels in \textsc{MathOlympiadEval}.

\subsection{Initial Evaluation of LLM Reasoning on \textsc{MathOlympiadEval}}
\label{sec:initial_eval_on_dataset}

Our initial evaluation using the \textsc{MathOlympiadEval} dataset reveals a critical phenomenon in current LLMs: models trained with outcome-based rewards frequently arrive at correct final answers through flawed or unsound reasoning processes. Table~\ref{tab:reasoning_analysis_combined_146} presents a unified analysis comparing answer correctness with human-verified reasoning correctness for 146 answer-finding problems from \textsc{MathOlympiadEval}.

\begin{table}[htbp]
\centering
\resizebox{0.45\textwidth}{!}{%
\begin{tabular}{lcccc}
\toprule
\textbf{Metric} & \textbf{CMO} & \textbf{IMO} & \textbf{OpenMathReasoning} & \textbf{Overall} \\
\midrule
Total Problems (Answer-Finding) & 59 & 31 & 56 & 146 \\
Answer Correctness (\%) & 88.1 & 41.9 & 92.9 & 80.1 \\
Human Correctness (Reasoning) (\%) & 47.5 & 0.0 & 53.6 & 39.7 \\
\midrule 
Correct Answers (Count) & 52 & 13 & 52 & 117 \\
Sound Reasoning Among Correct Answers (\%) & 53.8 & 0.0 & 57.7 & 49.6 \\
\bottomrule
\end{tabular}%
}
\caption{Unified analysis of answer correctness versus human-verified reasoning correctness, and the rate of sound reasoning among correct answers for the initial 146 answer-finding problems.}
\label{tab:reasoning_analysis_combined_146}
\end{table}

As Table~\ref{tab:reasoning_analysis_combined_146} illustrates, LLMs achieve a high overall answer correctness across these problems. However, the human-verified correctness of the underlying reasoning processes is substantially lower. This stark discrepancy indicates that models often employ flawed or incomplete reasoning even when producing the correct final answer.
This issue is particularly striking for the challenging IMO problems within \textsc{MathOlympiadEval}. While models achieve a notable answer correctness, none of these correct answers stem from sound reasoning. For CMO-Preliminary and OpenMathReasoning problems, reasoning correctness rates are higher, yet a substantial portion of correctly answered problems still lack valid supporting reasoning steps.

\section{ParaStepVerifier Methodology}

We introduce \textbf{ParaStepVerifier}, a novel automated methodology for evaluating mathematical reasoning. It verifies the logical soundness of each step in a proposed solution. This approach uses Large Language Models (LLMs) as verification agents, enabling detailed, interpretable, and potentially more accurate assessments than holistic or outcome-only methods.

\subsection{Step-by-Step Verification Task}

The core task is to assess the correctness of a candidate solution $S$ for a given mathematical problem $P$. 
We treat $S$ as an ordered sequence of reasoning steps $S = \{s_1, s_2, \ldots, s_n\}$. 
The entire solution's validity depends on the correctness of each step. 

\subsubsection{Verification Context and Objective}
Each reasoning step $s_i \in S$ is evaluated within a specific context. 
The primary verification context $C_i$ for step $s_i$ includes the problem statement $P$ and the history of preceding valid steps $H_{i-1} = \{s_1, s_2, \ldots, s_{i-1}\}$.
The objective is to determine if $s_i$ logically follows from this established context $(P, H_{i-1})$. 

\subsubsection{Error Typology}
A step $s_i$ is incorrect if it contains one or more reasoning errors. 
Section \ref{sec:dataset_construction_annotation_detailed} details the set of key error types $\mathcal{E}$ under the Annotation Process. 
Notably, ``Calculation Error'' is excluded from ParaStepVerifier's verification prompt, as function calls can often address such errors. 
Furthermore, our dataset defines ``Calculation Error'' to include computational mistakes within proof steps, not just in answer-finding problems. 

\subsubsection{Formal Verification Definition}
Let $m = |\mathcal{E}|$ be the number of error types. For each error type $j \in \{1, \ldots, m\}$, an error detection function $e_j(P, H_{i-1}, s_i)$ returns 1 if error type $j$ is present in deriving $s_i$ from its context, and 0 otherwise. 
The verification function $V$ for step $s_i$ determines its correctness:
\begin{equation}
V(s_i | P, H_{i-1}) = \begin{cases} 1 & \text{if } \sum\limits_{j=1}^{m} e_j(P, H_{i-1}, s_i) = 0 \\ 0 & \text{otherwise} \end{cases}
\label{eq:step_verification_revised}
\end{equation}
A value of 1 indicates a correct step (sum of errors is 0), while 0 indicates at least one error.
This can be expressed compactly as:
\begin{equation}
V(s_i | P, H_{i-1}) = \prod_{j=1}^{m} (1 - e_j(P, H_{i-1}, s_i))
\label{eq:step_verification_compact_revised} 
\end{equation}
The overall solution $S$ is correct if and only if all its steps are correct:
\begin{equation}
\text{Correctness}(S|P) = \prod_{i=1}^{n} V(s_i | P, H_{i-1})
\label{eq:solution_correctness_revised} 
\end{equation}

\subsection{LLM-based Implementation: The ParaStepVerifier Agent}

We implement the step-by-step verification using an LLM-based agent. 
This agent analyzes each step within its context to identify errors and provide a judgment. 

\subsubsection{Input Preparation: Solution Decomposition}
Our system automatically segments mathematical solutions into discrete steps, leveraging the inherent step-by-step structure of modern reasoning models. 
A solution $S$ is decomposed into an ordered sequence of reasoning steps $S = \{s_1, s_2, \dots, s_n\}$, where each $s_i$ is a coherent unit of mathematical reasoning. 
To ensure each step is substantial enough for evaluation, adjacent steps $s_k$ and $s_{k+1}$ are concatenated if $s_k$'s length is below a threshold $\theta$, empirically set to 12 tokens. 

\subsubsection{Input Preparation: Context Construction for LLM}
To verify a step $s_i$, the LLM agent receives a carefully constructed context. 
This context includes the original problem $P$, preceding steps $H_{i-1} = \{s_1, s_2, \ldots, s_{i-1}\}$, the current step $s_i$, and any future steps $F_{i+1} = \{s_{i+1}, \ldots, s_n\}$. 
The full context for verifying $s_i$ is the tuple $(P, H_{i-1}, s_i, F_{i+1})$, termed the Analysis Context. 
While future steps $F_{i+1}$ help the LLM understand the solution's trajectory, the agent's primary task is to judge $s_i$'s correctness based strictly on its logical derivability from the Verification Context $(P, H_{i-1})$, guided by our prompt structure (Section \ref{sec:appendix_prompt_templates}).

\subsubsection{Adaptive Verification Strategy}
ParaStepVerifier employs an adaptive strategy based on the number of solution steps to enhance evaluation efficiency and accuracy. 
For concise solutions with a single reasoning step, the system uses a direct approach combining LLM assessment with targeted error detection for rapid validation. 
For complex, multi-step solutions, ParaStepVerifier activates its full step-by-step verification mechanism. 
It meticulously analyzes each step $s_i$, using the sequence of previously validated steps $H_{i-1}$ and subsequent steps $F_{i+1}$ as context. 

\subsection{System Architecture and Parallelization}

To efficiently process long solutions, ParaStepVerifier is implemented within a parallel processing architecture.
Key components are:
\noindent\textbf{Solution Decomposition Module}: Implements step segmentation and concatenation. \textbf{Task Queue}: Holds verification tasks, each corresponding to a single step $s_i$ and its context. \textbf{Parallel Verification Workers}: A pool of processes, managed by tools like \texttt{concurrent.futures.ProcessPoolExecutor}. Each worker takes a task, constructs a prompt, queries the LLM, and parses the judgment. \textbf{Resource Management}: Adaptively manages worker numbers based on available resources. Batch processing is used with the LLM API for efficiency. \textbf{Result Aggregation}: Collects True/False judgments for all steps to determine overall solution correctness via Equation \ref{eq:solution_correctness_revised}.
This parallel architecture significantly reduces evaluation time for complex mathematical solutions.

\section{Experiment}

\subsection{Settings}

\paragraph{Dataset and Baselines}
We evaluate our approach on our datadet \textsc{MathOlympiadEval}.We evaluate the performance of several models that have shown significant capabilities in mathematical reasoning tasks, including Gemini-2.0-flash, QwQ-32B, o1-mini, o3-mini, Qwen3-32B, o1, and Gemini-2.5-Pro. These models serve as baseline verification agents for mathematical proofs or solutions.

\paragraph{Metrics}
To evaluate the performance of our verification model in identifying flawed mathematical solutions or proofs, our primary evaluation metric is the F1 score. We frame this task as a binary classification problem where the "positive" class signifies that a given solution is incorrect.

Let $\mathcal{S}_{\text{incorrect}}^{\text{GT}}$ denote the set of solutions labeled as incorrect in the ground truth, and let $\mathcal{S}_{\text{incorrect}}^{\text{Pred}}$ denote the set of solutions identified as incorrect by our model. True Positives (TP), False Positives (FP), False Negatives (FN), Precision, Recall, and F1 score are defined as follows:
\begin{align}
    &\text{TP} = |\mathcal{S}_{\text{incorrect}}^{\text{GT}} \cap \mathcal{S}_{\text{incorrect}}^{\text{Pred}}|, \\
    &\text{FP} = |\mathcal{S}_{\text{incorrect}}^{\text{Pred}} \setminus \mathcal{S}_{\text{incorrect}}^{\text{GT}}|, \\
    &\text{FN} = |\mathcal{S}_{\text{incorrect}}^{\text{GT}} \setminus \mathcal{S}_{\text{incorrect}}^{\text{Pred}}|, \\
    &\text{Precision} = \frac{\text{TP}}{\text{TP} + \text{FP}}, \\
    &\text{Recall} = \frac{\text{TP}}{\text{TP} + \text{FN}}, \\
    &\text{F1} = 2 \times \frac{\text{Precision} \times \text{Recall}}{\text{Precision} + \text{Recall}}. \label{eq:f1_score_full}
\end{align}

\subsection{Results}
\label{sec:results}

This section details the empirical performance of ParaStepVerifier. We first demonstrate its overall effectiveness in enhancing the accuracy of mathematical reasoning verification and its comparative cost-efficiency. We then analyze its robustness in evaluating solutions with extended reasoning chains.

\subsubsection{Overall Verification Performance and Cost-Effectiveness}

Our evaluations, summarized in Table~\ref{tab:results_in_column}, reveal that ParaStepVerifier substantially improves F1 scores for identifying incorrect mathematical solutions when compared to standard LLM-as-a-judge baselines across various base models. For example, even a capable base model like Gemini-2.0-flash sees its Overall F1 score improve from 75.89\% to 83.39\% when augmented with ParaStepVerifier. This highlights ParaStepVerifier's ability to guide base models to more accurate evaluations through its structured, step-by-step verification process, with consistent performance improvements observed across diverse mathematical sub-datasets.

ParaStepVerifier enables base models to achieve evaluation capabilities that can surpass even stronger, state-of-the-art models, such as o1, while ParaStepVerifier with the base models is much more cost-effective than the stronger models.
Here the stronger models are used in a simpler LLM-as-a-judge capacity. As detailed in Table~\ref{tab:results_in_column}, ParaStepVerifier combined with Gemini-2.0-flash achieve the best performance (Overall F1: 83.39\%) and is also the most cost-effective in the agent solutions. It slightly outperforms Gemini2.5Pro, (Overall F1: 83.26\%), the best LLM-as-a-judge solution at an estimated cost of \$0.97, considerably less than the \$9.32 for Gemini2.5Pro or \$4.12 for o3-mini when used as LLM-as-a-judge, despite the latter two having lower F1 scores. This indicates that ParaStepVerifier offers an effective pathway to significantly improve accuracy in mathematical reasoning evaluation, often with greater cost-efficiency.

While ParaStepVerifier demonstrates significant improvements with the tested models, we currently do not present results for ParaStepVerifier integrated with the most powerful LLMs listed under the LLM-as-a-judge category, such as o1 or Gemini2.5Pro. This is primarily due to the substantial computational costs and resource demands associated with the extensive, step-by-step verification process inherent to ParaStepVerifier. Each solution verification using this method involves multiple LLM invocations, and applying this to the largest models would lead to considerably higher expenses and longer processing times, which were beyond the scope of the current experimental setup. Future work may explore these configurations as resources permit and model API costs evolve.

% Table 1: Evaluation results
\begin{table}[h!]
  \centering
  \resizebox{\columnwidth}{!}{%
    \begin{tabular}{@{}lccccc@{}}
    \toprule
    \textbf{Model} & \multicolumn{4}{c}{\textbf{F1 Score (\%)}} & \textbf{Cost (\$)} \\
    \cmidrule(lr){2-5}
    & \textbf{CMO} & \textbf{IMO} & \textbf{OpenMathReasoning} & \textbf{Overall} & \\
    \midrule
    \multicolumn{6}{@{}c@{}}{\textbf{LLM-as-a-judge}} \\
    \cmidrule(lr){1-6}
    QwQ-32B           & 43.33 & 65.22 & 37.84 & 52.91 & 0.84 \\ 
    Qwen3-32B         & 51.61 & 78.43 & 40.00 & 62.75 & 1.04 \\ 
    Gemini-2.0-flash  & 57.14 & 94.92 & 62.75 & 75.89 & 0.37 \\ 
    o1-mini           & 48.57 & 77.23 & 41.86 & 60.75 & 4.10 \\ 
    o3-mini           & 67.47 & 95.80 & 69.09 & 80.93 & 4.12 \\ 
    o1                & 65.71 & 94.02 & 20.00 & 74.65 & 56.21 \\ 
    Gemini2.5Pro      & 70.89 & 97.52 & 70.37 & 83.26 & 9.32 \\ 
    \midrule
    \multicolumn{6}{@{}c@{}}{\textbf{ParaStepVerifier}} \\
    \cmidrule(lr){1-6}
    ParaStepVerifier\_o1-mini        & 68.89 & 77.23 & 64.00 & 71.37 & 10.64 \\ 
    ParaStepVerifier\_QwQ-32B        & 61.73 & 90.27 & 58.82 & 74.29 & 2.18 \\ 
    ParaStepVerifier\_Qwen3-32B      & 72.09 & 94.92 & 72.73 & 82.63 & 2.71 \\ 
    ParaStepVerifier\_Gemini-2.0-flash & 69.57 & 97.52 & 75.86 & 83.39 & 0.97 \\ 
    \bottomrule
    \end{tabular}%
  }
  \caption{Evaluation results on \MathOlympiadEval. We report F1 scores (\%) and estimated evaluation costs (\$) on the \MathOlympiadEval dataset.}
  \label{tab:results_in_column}
\end{table}

\subsubsection{Performance on Solutions with Extended Reasoning Chains}

To assess ParaStepVerifier's effectiveness on more complex problems, we compare its performance against the LLM+EC (LLM-as-a-judge with Error Classification) strategy for solutions with extended reasoning chains. Table~\ref{tab:perf_by_step_length_multirow_relative_delta} details this comparison, highlighting the relative F1 score improvements achieved by ParaStepVerifier over LLM+EC.

\begin{table}[htbp]
    \centering
    \resizebox{0.45\textwidth}{!}{
        \begin{tabular}{llcc}
            \toprule
            \textbf{Base Model} & \textbf{Verification} & \textbf{F1 (\%)} & \textbf{F1 (\%)} \\
            & \textbf{Strategy} & \textbf{(Step >7)} & \textbf{(Step >9)} \\
            \midrule
            \multirow{3}{*}{QwQ-32B} & LLM+EC & 72.22 & 66.67 \\
            & ParaStepVerifier & 84.85 & 85.71 \\
            & \textit{Rel. Improv. (\% $\Delta$)} & \textbf{\textcolor{ForestGreen}{+17.49\%}} & \textbf{\textcolor{ForestGreen}{+28.56\%}} \\
            \cmidrule(lr){1-4}
            \multirow{3}{*}{o1-mini} & LLM+EC & 60.61 & 44.44 \\
            & ParaStepVerifier & 72.22 & 66.67 \\
            & \textit{Rel. Improv. (\% $\Delta$)} & \textbf{\textcolor{ForestGreen}{+19.15\%}} & \textbf{\textcolor{ForestGreen}{+50.02\%}} \\
            \cmidrule(lr){1-4}
            \multirow{3}{*}{Qwen3-32B} & LLM+EC & 80.00 & 85.71 \\
            & ParaStepVerifier & 87.18 & 88.89 \\
            & \textit{Rel. Improv. (\% $\Delta$)} & \textbf{\textcolor{ForestGreen}{+8.98\%}} & \textbf{\textcolor{ForestGreen}{+3.71\%}} \\
            \cmidrule(lr){1-4}
            \multirow{3}{*}{Gemini-2.0-flash} & LLM+EC & 83.33 & 66.67 \\
            & ParaStepVerifier & 85.00 & 85.71 \\
            & \textit{Rel. Improv. (\% $\Delta$)} & \textbf{\textcolor{ForestGreen}{+2.00\%}} & \textbf{\textcolor{ForestGreen}{+28.56\%}} \\
            \bottomrule
        \end{tabular}%
    }
    \caption{F1 performance comparison between verification strategies on extended reasoning chains. A dedicated row for each model shows the \textit{Relative Improvement (Rel. Improv. \% $\Delta$)} in F1 score (\textbf{\textcolor{ForestGreen}{bold green values}}) achieved by ParaStepVerifier over LLM+EC.}
    \label{tab:perf_by_step_length_multirow_relative_delta}
\end{table}

The results in Table~\ref{tab:perf_by_step_length_multirow_relative_delta} demonstrate that ParaStepVerifier consistently yields substantial relative F1 score improvements as solution complexity increases (number of steps > 7 and > 9). For example, with QwQ-32B as the base model, ParaStepVerifier achieves a relative F1 score improvement of 17.49\% for solutions with more than 7 steps, and a significant 28.56\% for those with more than 9 steps, compared to LLM+EC. Similar impactful relative gains are observed across other base models; o1-mini records a remarkable relative improvement of 50.02\% for the most complex problems (>9 steps). Even for stronger baselines like Qwen3-32B and Gemini-2.0-flash, ParaStepVerifier provides clear advantages, with Gemini-2.0-flash showing a 28.56\% relative F1 increase on solutions with >9 steps.

This consistent trend of significant positive relative improvements, as detailed in the \textit{Rel. Improv. (\% $\Delta$)} rows for each model in Table~\ref{tab:perf_by_step_length_multirow_relative_delta}, underscores ParaStepVerifier's enhanced capability in accurately verifying solutions that demand extensive, multi-step reasoning.

% Table 2: Performance by step length (with Relative Delta)

\section{Ablations}
\label{sec:ablations} 

In this section, we conduct ablation studies to evaluate crucial aspects of our ParaStepVerifier pipeline.
First, in Section~\ref{sec:Error Classification and Step-by-Step Verification}, we analyze the individual and combined impacts of Error Classification and Step-by-Step Verification, which are core mechanisms for assessing the correctness of mathematical reasoning.
Subsequently, in Section~\ref{sec:effect_parallel_verification}, we examine the effectiveness of Parallel Verification in improving processing efficiency compared to sequential approaches.

\subsection{Error Classification and Step-by-Step Verification}
\label{sec:Error Classification and Step-by-Step Verification}

\begin{table}[htbp]
  \centering
  \resizebox{\linewidth}{!}{%
  \begin{tabular}{@{}llcccc@{}}
    \toprule
    \textbf{Base Model} & \textbf{Verification Strategy} & \textbf{CMO} & \textbf{IMO} & \textbf{OpenMathReasoning} & \textbf{Overall} \\
    \midrule
    \multirow{4}{*}{QwQ-32B} & LLM-as-a-judge & 43.33 & 65.22 & 37.84 & 52.91 \\
                              & ParaStepVerifier w/o EC & 47.76   & 70.83   & 42.86   & 57.56   \\
                              & LLM+EC & 55.07   & 81.90   & 50.00   & 66.67   \\
                              & ParaStepVerifier & \textbf{61.73} & \textbf{90.27} & \textbf{58.82} & \textbf{74.29} \\
    \midrule
    \multirow{4}{*}{o1-mini} & LLM-as-a-judge & 48.57 & 77.23 & 41.86 & 60.75 \\
                              & ParaStepVerifier w/o EC & 58.54   & \textbf{86.24}   & 44.44   & 68.64   \\
                              & LLM+EC & 58.97   & 70.83   & 63.83   & 65.16   \\
                              & ParaStepVerifier & \textbf{68.89} & 77.23 & \textbf{64.00} & \textbf{71.37} \\
    \midrule
    \multirow{4}{*}{Qwen3-32B} & LLM-as-a-judge & 51.61 & 78.43 & 40.00 & 62.75 \\
                              & ParaStepVerifier w/o EC & 58.33   & 88.29   & 52.17   & 71.62   \\
                                & LLM+EC & 65.85   & 90.27   & 65.31   & 77.05   \\
                                & ParaStepVerifier & \textbf{72.09} & \textbf{94.92} & \textbf{72.73} & \textbf{82.63} \\
    \midrule
    \multirow{4}{*}{Gemini-2.0-flash} & LLM-as-a-judge & 57.14 & 94.92 & 62.75 & 75.89 \\
                                      & ParaStepVerifier w/o EC & 59.26   & 90.27   & 60.38   & 73.68   \\
                                      & LLM+EC & \textbf{69.57}   & \textbf{97.52}   & 73.33   & 82.78   \\
                                      & ParaStepVerifier & \textbf{69.57} & \textbf{97.52} & \textbf{75.86} & \textbf{83.39} \\
    \bottomrule
  \end{tabular}%
  } 
    \caption{Ablation study on the impact of error classification and step-by-step verification. F1 scores are reported for different verification strategies across our mathematical reasoning datasets \textsc{MathOlympiadEval}.}
    \label{tab:combined_ablation_f1_reordered_bold} 
\end{table}

This section analyzes the impact of Error Classification (EC) and Step-by-Step Verification, two key components of ParaStepVerifier. Table~\ref{tab:combined_ablation_f1_reordered_bold} presents F1 scores demonstrating their individual and combined contributions.

\paragraph{Error Classification}
The Error Classification (EC) module consistently enhances verification performance. First, adding EC to the standard LLM-as-a-judge approach to form LLM+EC improves Overall F1 scores for all base models. For example, QwQ-32B and Qwen3-32B show notable gains. Second, EC also boosts our step-by-step method: the full \textbf{ParaStepVerifier} with EC outperforms ParaStepVerifier w/o EC across all models. These results demonstrate that EC refines the verification process by enabling more precise judgments.

\paragraph{Step-by-Step Verification}
Step-by-Step Verification also provides clear benefits over holistic evaluation. Even without EC, ParaStepVerifier w/o EC generally achieves higher Overall F1 scores than the LLM-as-a-judge baseline, as seen with models like QwQ-32B and o1-mini. This highlights the inherent advantage of decomposing the verification task. When EC is integrated, the superiority of the step-by-step approach becomes more pronounced. The full \textbf{ParaStepVerifier} consistently surpasses LLM+EC, further improving Overall F1 scores for all base models.

\subsection{Parallel Verification}
\label{sec:effect_parallel_verification}

To evaluate the efficiency gains from our parallel processing architecture, we analyzed the runtime of different verification strategies, detailed in Table \ref{tab_effect_parallel_verification_transposed}. This table compares ParaStepVerifier, its sequential counterpart SeqStepVerifier, and the LLM-as-a-judge baseline.

\begin{table}[htbp]
    \centering
    \resizebox{0.45\textwidth}{!}{%
        \begin{tabular}{lcccc}
            \toprule
            \textbf{Verification Strategy} & \textbf{QwQ-32B} & \textbf{Qwen3-32B} & \textbf{o1-mini} & \textbf{Gemini-2.0-flash} \\
            \midrule
            LLM-as-a-judge & 8.04 & 6.61 & 1.22 & 0.40 \\
            SeqStepVerifier (Sequential) & 23.78 & 19.36 & 4.05 & 1.11 \\
            ParaStepVerifier (Parallel)   & 9.88 & 7.48 & 1.88 & 0.51 \\
            \bottomrule
        \end{tabular}%
    }
    \caption{Runtime analysis of parallel verification strategies (hours)}
    \label{tab_effect_parallel_verification_transposed}
\end{table}

The data indicate that parallelization significantly reduces verification time. For all tested models, ParaStepVerifier substantially decreased runtime compared to SeqStepVerifier, achieving, for example, a speedup of over 2.4 times on QwQ-32B. This consistent acceleration across models demonstrates the effectiveness of ParaStepVerifier's parallel architecture, crucial for the timely processing of complex mathematical solutions.

Although ParaStepVerifier inherently incurs a higher computational cost than the LLM-as-a-judge baseline due to its detailed step-by-step analysis, parallelization keeps its runtime comparable to this baseline. Therefore, ParaStepVerifier's parallel strategy effectively mitigates the computational overhead associated with fine-grained, step-by-step analysis, offering a practical and scalable solution for robust mathematical reasoning evaluation while achieving a significantly more thorough assessment than LLM-as-a-judge.

\section{Conclusion}
\label{sec:conclusion}

This paper investigated a significant challenge in the advancement of artificial intelligence: the tendency for Large Language Models to arrive at correct answers in mathematical reasoning tasks through processes that are fundamentally flawed. This issue is often masked by evaluation methods that prioritize final outcomes over the integrity of the reasoning steps. Our work empirically confirmed a critical disconnect between superficial answer accuracy and genuine procedural soundness in contemporary LLMs, frequently attributing this to "reward hacking," where models exploit evaluation metrics without developing true understanding, leading to prevalent errors such as unjustified guessing or significant logical deficiencies.

To address this, we developed and validated \textbf{ParaStepVerifier}, a novel methodology employing an agent-based system for meticulous, step-by-step verification of mathematical solutions. Through comprehensive evaluations utilizing our newly constructed \textbf{MathOlympiadEval} dataset—a diverse collection of challenging mathematical problems with detailed human annotations of reasoning correctness—we demonstrated that ParaStepVerifier substantially elevates the accuracy and reliability of mathematical reasoning assessment. Our findings show consistent improvements in identifying incorrect solutions when ParaStepVerifier is applied, surpassing traditional holistic evaluation approaches, particularly when dealing with complex problems that necessitate extended chains of reasoning. This enhanced performance was often achieved with greater cost-effectiveness. Further analyses of ParaStepVerifier's components confirmed the distinct benefits of its step-by-step analysis and integrated error classification capabilities.

ParaStepVerifier offers a more robust and transparent framework for scrutinizing the reasoning pathways of LLMs. It serves as a valuable tool for more dependable model evaluation, for guiding the creation of higher-quality training datasets, and ultimately, for fostering the development of artificial intelligence systems with more reliable and verifiable mathematical reasoning skills. While this research establishes a strong foundation, future explorations could involve integrating ParaStepVerifier with the most advanced large language models, as computational practicalities evolve, and adapting its principles to assess complex reasoning in other specialized domains.

\section*{Limitations}

While ParaStepVerifier offers a significant step towards more reliable mathematical reasoning evaluation, we acknowledge certain limitations.

\paragraph{Generalizability Across Domains and Reasoning Types.}
Our findings are primarily based on the \textsc{MathOlympiadEval} dataset, which emphasizes structured mathematical problems, including answer-finding and proof-based tasks. While its performance on these mathematical tasks is promising, its effectiveness on problems from substantially different mathematical fields (e.g., highly abstract topology, quantum field theory derivations) or those with unique structural presentations requires further validation. Moreover, given ParaStepVerifier's capability in verifying logical steps in mathematical proofs, its potential to generalize to broader textual reasoning domains—such as evaluating the coherence of arguments in legal texts or scientific papers—is an intriguing area for future research but remains unconfirmed by the current study.

\paragraph{Dependence on Base LLM Capabilities.}
The effectiveness of ParaStepVerifier is inherently linked to the reasoning and verification capabilities of the LLM it employs as a verifier. Although our framework guides the LLM to more accurate assessments, the inherent limitations or biases of the chosen base model can still influence the verification outcome, especially for novel or extremely complex reasoning. Full-scale testing of ParaStepVerifier using the largest state-of-the-art LLMs as the verifier agent was constrained by current computational costs.

\bibliography{custom}

\appendix

\onecolumn
\section{Dataset Details}

This section details the construction and key characteristics of the \textsc{MathOlympiadEval} dataset, a benchmark for evaluating mathematical reasoning in Large Language Models (LLMs).

\subsection{Dataset Composition and Problem Types}
\textsc{MathOlympiadEval} contains 204 mathematical problems sourced from: the China Mathematical Olympiad (CMO, 86 problems), the International Mathematical Olympiad (IMO, 62 problems), and OpenMathReasoning (56 problems). The dataset is diverse, comprising 58 proof-based problems and 146 problems requiring a final numerical or symbolic answer. Problems span a range of difficulties, with IMO problems representing the most challenging tier.

\subsection{Solution Generation and Annotation}
\begin{figure}[t]
    \centering
    \includegraphics[width=0.7\columnwidth]{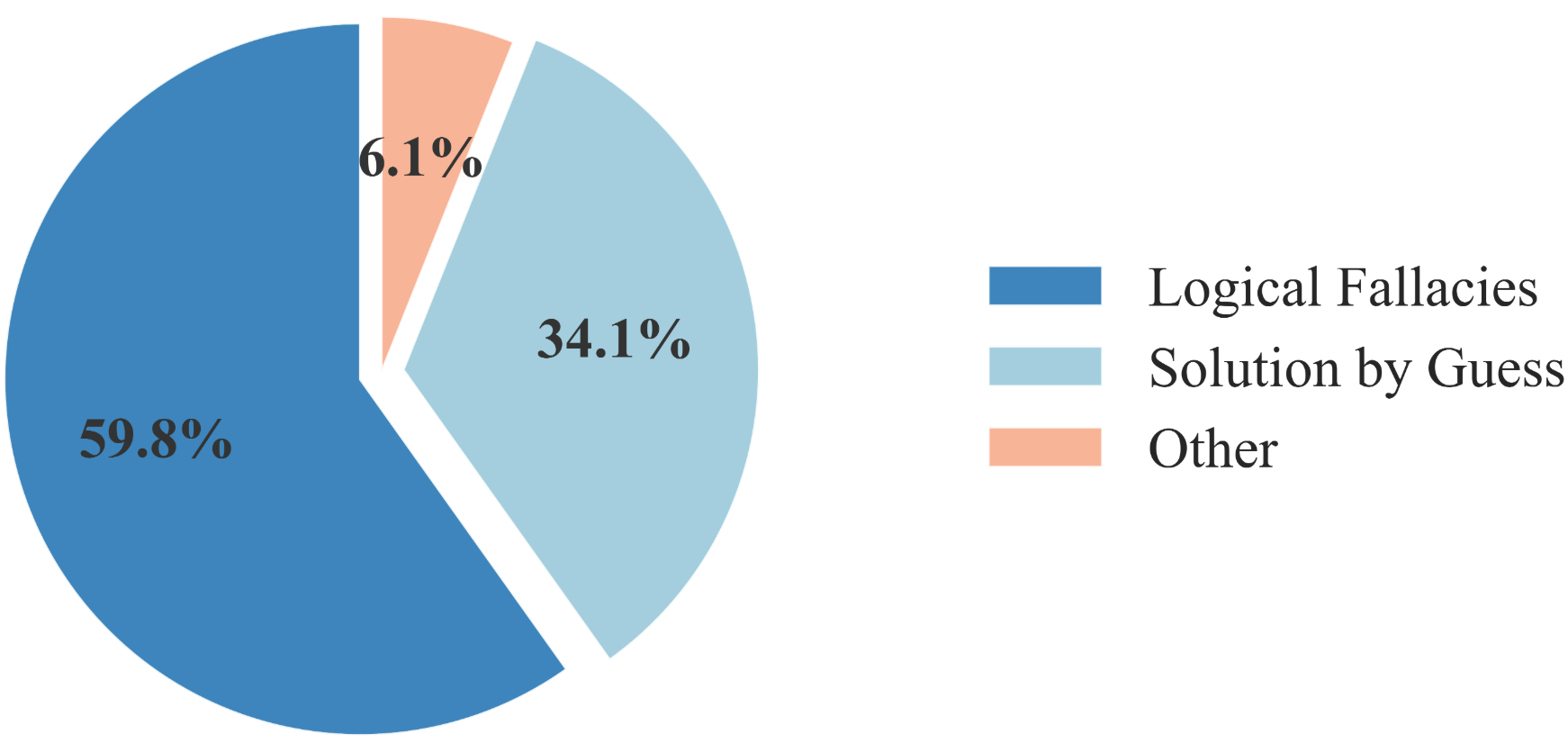} 
    \caption{Distribution of primary error types identified in LLM-generated solutions with flawed reasoning. Logical fallacies and guessing constitute the majority of errors.}
    \label{fig:error_distribution}
\end{figure}

We generated solutions using open-source LLMs trained via outcome-based reward strategies. Prompts were designed to elicit complete, step-by-step reasoning. Each generated solution underwent a rigorous dual-evaluation process:
\begin{itemize}
    \item \textbf{Answer Correctness}: Automatic verification of the final answer against a reference solution.
    \item \textbf{Reasoning Correctness}: Manual expert assessment of the logical soundness of the entire reasoning process.
\end{itemize}
To ensure high-quality annotations, a three-round iterative expert review process was implemented. Solutions with flawed reasoning were categorized by error type. As illustrated in Figure~\ref{fig:error_distribution}, logical fallacies (59.8\%) and solutions derived by guessing (34.1\%) are the most prevalent error categories, underscoring systemic weaknesses in current LLM reasoning capabilities.

\subsection{Reasoning Complexity and Structure}
\begin{figure}[t]
  \centering
  \includegraphics[width=0.9\columnwidth]{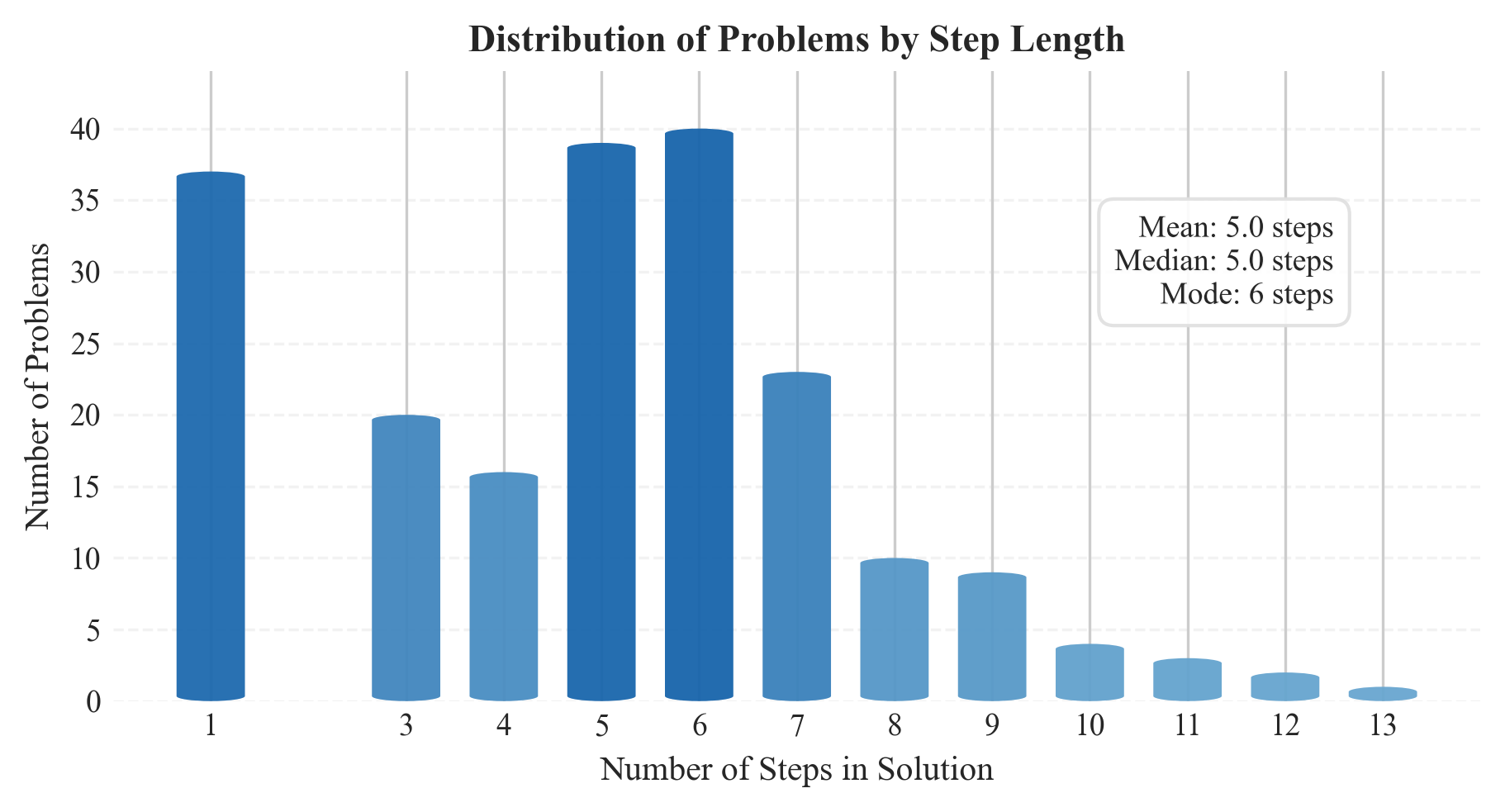} 
  \caption{Distribution of problems by the number of reasoning steps in their solutions. The x-axis indicates the number of steps, and the y-axis shows the count of problems. The distribution highlights that most solutions involve a moderate number of steps, suitable for detailed reasoning analysis.}
  \label{fig:step_distribution}
\end{figure}

An analysis of solution complexity (Figure~\ref{fig:step_distribution}) reveals an average of $5.04$ reasoning steps per problem, with a median of $5$ and a mode of $6$ steps. The number of steps ranges from $1$ to $13$. Notably, approximately $75\%$ of solutions require $7$ or fewer steps. This complexity level is substantial enough to challenge current models yet remains manageable for fine-grained human analysis of the reasoning process.

\subsection{Value for LLM Reasoning Research}
\textsc{MathOlympiadEval} provides a crucial benchmark with fine-grained human annotations specifically designed for advancing research in LLM mathematical reasoning. Its primary contribution lies in facilitating the evaluation of the \textit{reasoning process itself}, rather than solely focusing on final answer correctness. This distinction is vital for uncovering instances of "correct answers via flawed reasoning," a phenomenon that can mask true model capabilities and hinder progress. By offering insights into these deficiencies, the dataset supports the development of strategies to mitigate issues like reward hacking and ultimately build more reliable and interpretable mathematical reasoning systems.

\subsection{Dataset Format}
\textsc{MathOlympiadEval} consists of 204 records. Each record includes 13 distinct fields: problem description, model-generated solution, detailed reasoning content, binary evaluation of answer correctness, human assessment of reasoning correctness, reference answer, problem type (proof-based or answer-finding), details of the LLM used for generation, problem source, primary error type (if applicable), and the total number of reasoning steps.

\section{Prompt Templates}
\label{sec:appendix_prompt_templates}

This section presents the prompt templates used for mathematical reasoning evaluation. Templates cover holistic assessment for baselines and step-by-step verification for our ParaStepVerifier, with variations including error classification to support ablation studies.

\subsection{LLM-as-a-judge Prompt Templates}

LLM-as-a-judge, our baseline, performs a one-shot assessment of an entire solution's correctness.
\begin{figure*}[htbp]
    \centering
    \includegraphics[width=0.8\linewidth]{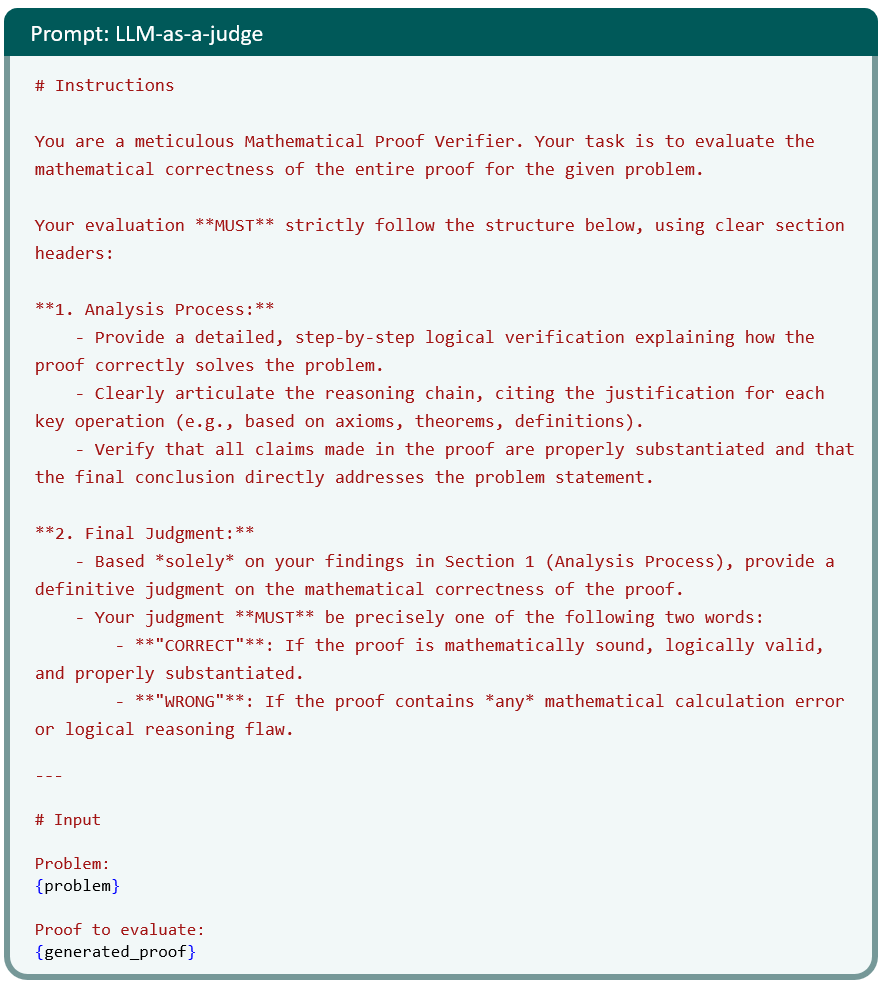}
    \caption{Basic LLM-as-a-judge prompt for holistic solution evaluation.}
    \label{fig:prompt_llm_as_a_judge}
\end{figure*}

\begin{figure*}[htbp]
    \centering
    \includegraphics[width=0.8\linewidth]{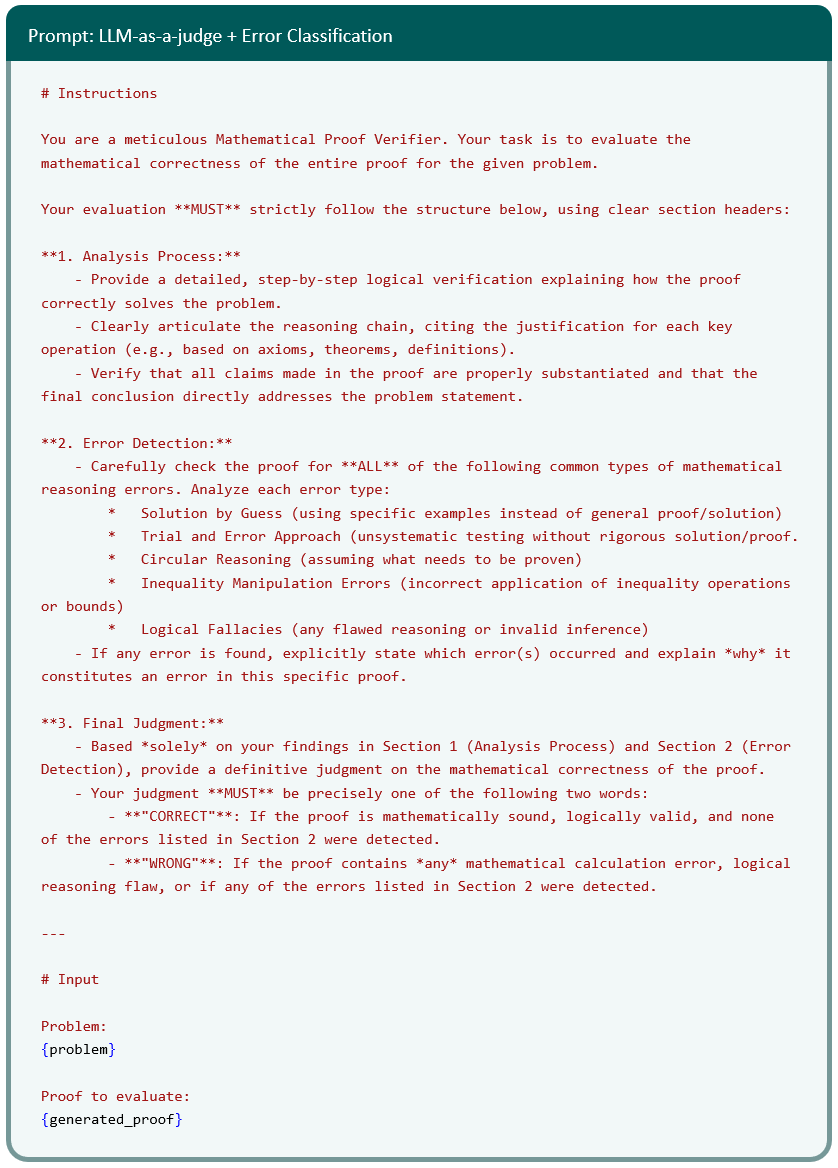} 
    \caption{LLM-as-a-judge prompt with an added error classification requirement.}
    \label{fig:prompt_llm_as_a_judge_with_ec}
\end{figure*}

\subsection{ParaStepVerifier Prompt Templates}

ParaStepVerifier, our proposed method, performs fine-grained, step-by-step verification by assessing each solution step independently.
\begin{figure*}[htbp] 
    \centering
    \includegraphics[width=0.8\linewidth]{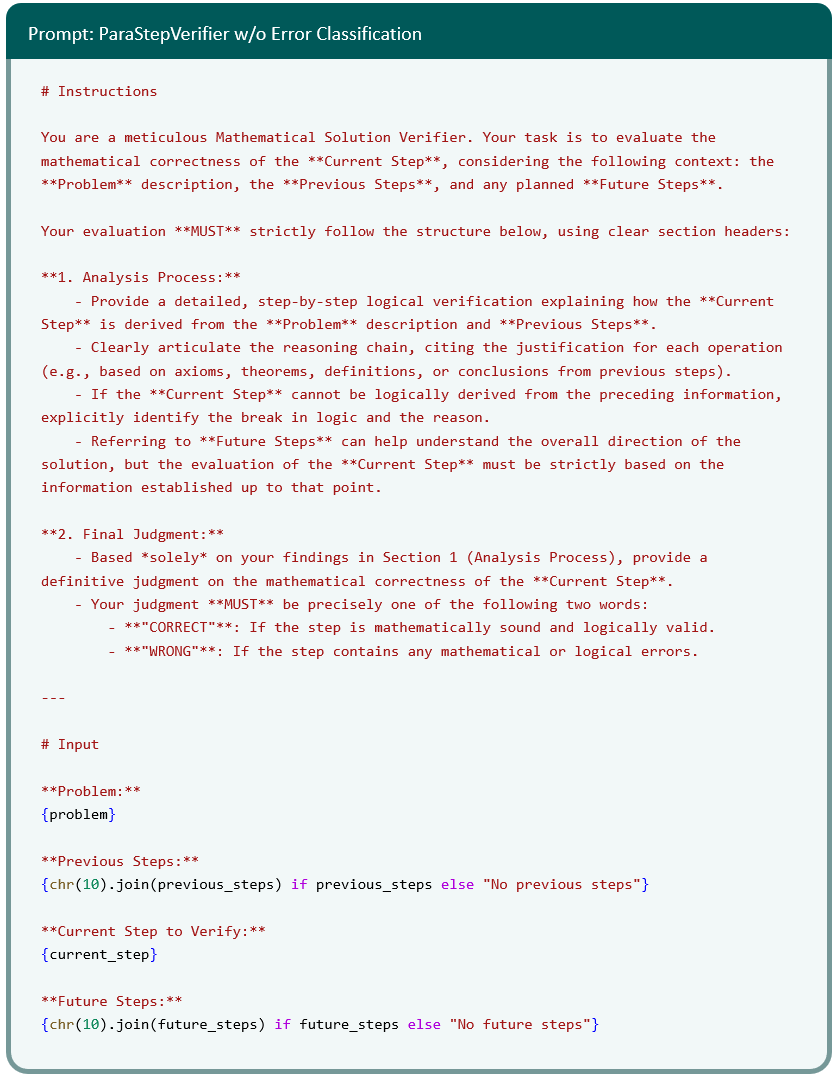} 
    \caption{ParaStepVerifier prompt for step-wise evaluation without explicit error categories listed in the instruction.}
    \label{fig:prompt_parastepverifier_without_ec}
\end{figure*}

\begin{figure*}[htbp] 
    \centering
    \includegraphics[width=0.8\linewidth]{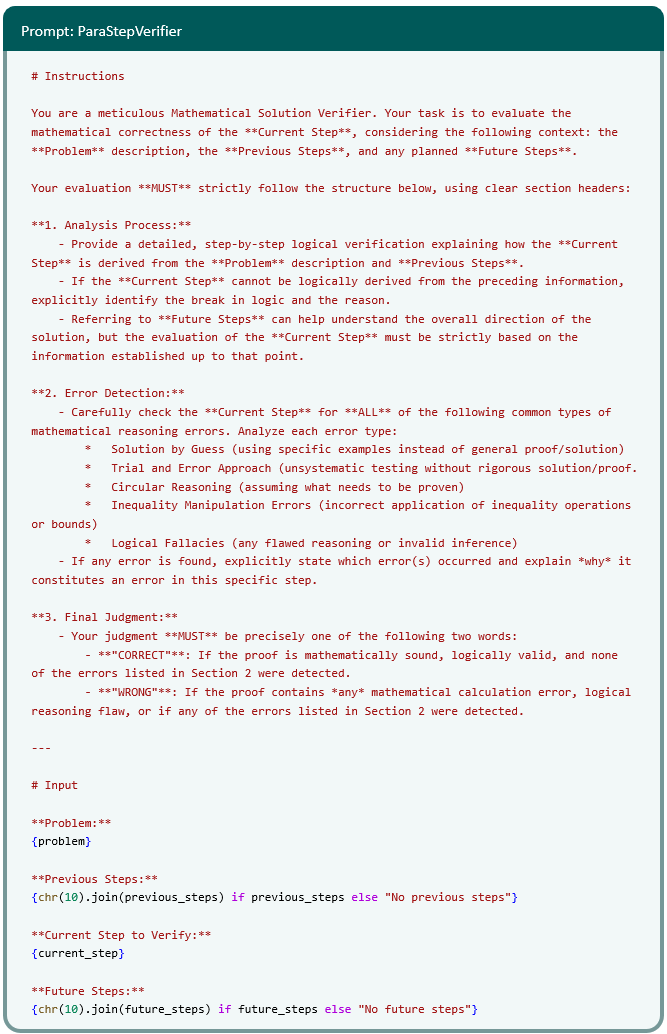} 
    \caption{ParaStepVerifier prompt with step-wise evaluation and explicit error classification.}
    \label{fig:prompt_parastepverifier_with_ec}
\end{figure*}

\section{Case Study}

To illustrate the core findings of our paper—that models often achieve correct final answers through flawed reasoning—this section presents a series of case studies drawn from the \textsc{MathOlympiadEval} dataset. These examples, sourced from IMO, CMO, and OpenMathReasoning, showcase various failure modes, including "Solution by Guess," "Calculation Errors," "Inequality Manipulation Errors," and other "Logical Fallacies." Each case serves as a clear example of reward hacking, highlighting the critical gap between outcome-based correctness and process-based soundness, and underscoring the need for more rigorous evaluation methods like ParaStepVerifier.

\subsection{OpenMathReasoning}

\begin{casebox}{OpenMathReasoning}{Solution by Guess}
\textbf{Problem Description:} Find all non-negative integers $m$ and $n$ such that $(2^n-1) \cdot (3^n-1) = m^2$.

\textbf{Error Category:} \textit{Solution by Guess}

\textbf{Reason for Error:} The solution relies on pattern-spotting from a few small examples ($n=0, 1, 2, ...$) and fails to provide a rigorous proof that no solutions exist for larger values of $n$. Through this flawed and incomplete derivation, it  arrives at the correct answer.
\tcblower
\textbf{Solution:}

To find all non-negative integers \( m \) and \( n \) such that \((2^n - 1)(3^n - 1) = m^2\), we will proceed as follows:

1. **Check small values of \( n \):**
   - For \( n = 0 \):
     \[
     (2^0 - 1)(3^0 - 1) = (1 - 1)(1 - 1) = 0 \cdot 0 = 0 = 0^2
     \]
     So, \( (m, n) = (0, 0) \) is a solution.

   - For \( n = 1 \):
     \[
     (2^1 - 1)(3^1 - 1) = (2 - 1)(3 - 1) = 1 \cdot 2 = 2
     \]
     Since 2 is not a perfect square, \( n = 1 \) is not a solution.

   - For \( n = 2 \):
     \[
     (2^2 - 1)(3^2 - 1) = (4 - 1)(9 - 1) = 3 \cdot 8 = 24
     \]
     Since 24 is not a perfect square, \( n = 2 \) is not a solution.

   - For \( n = 3 \):
     \[
     (2^3 - 1)(3^3 - 1) = (8 - 1)(27 - 1) = 7 \cdot 26 = 182
     \]
     Since 182 is not a perfect square, \( n = 3 \) is not a solution.

   - For \( n = 4 \):
     \[
     (2^4 - 1)(3^4 - 1) = (16 - 1)(81 - 1) = 15 \cdot 80 = 1200
     \]
     Since 1200 is not a perfect square, \( n = 4 \) is not a solution.

   - For \( n = 5 \):
     \[
     (2^5 - 1)(3^5 - 1) = (32 - 1)(243 - 1) = 31 \cdot 242 = 7502
     \]
     Since 7502 is not a perfect square, \( n = 5 \) is not a solution.

   - For \( n = 6 \):
     \[
     (2^6 - 1)(3^6 - 1) = (64 - 1)(729 - 1) = 63 \cdot 728 = 45864
     \]
     Since 45864 is not a perfect square, \( n = 6 \) is not a solution.

2. **Analyze the general case:**

   Let \( A = 2^n - 1 \) and \( B = 3^n - 1 \). Then \( A \cdot B = m^2 \). For \( A \cdot B \) to be a perfect square, \( A \) and \( B \) must share a common factor \( d \) such that \( d \) is a perfect square, and the remaining factors of \( A \) and \( B \) after dividing by \( d \) must also be perfect squares.

   - Compute the GCD of \( 2^n - 1 \) and \( 3^n - 1 \):
     \[
     \text{GCD}(2^n - 1, 3^n - 1)
     \]
     For \( n \geq 1 \), \( 2^n - 1 \) is odd and \( 3^n - 1 \) is even. Therefore, the GCD must be odd.

   - Check small values of \( n \) to see if the GCD is a perfect square:
     - For \( n = 4 \):
       \[
       \text{GCD}(2^4 - 1, 3^4 - 1) = \text{GCD}(15, 80) = 5
       \]
       Since 5 is not a perfect square, \( n = 4 \) is not a solution.

     - For \( n = 6 \):
       \[
       \text{GCD}(2^6 - 1, 3^6 - 1) = \text{GCD}(63, 728) = 7
       \]
       Since 7 is not a perfect square, \( n = 6 \) is not a solution.

     - For \( n = 8 \):
       \[
       \text{GCD}(2^8 - 1, 3^8 - 1) = \text{GCD}(255, 6560) = 5
       \]
       Since 5 is not a perfect square, \( n = 8 \) is not a solution.

     - For \( n = 10 \):
       \[
       \text{GCD}(2^{10} - 1, 3^{10} - 1) = \text{GCD}(1023, 59048) = 11
       \]
       Since 11 is not a perfect square, \( n = 10 \) is not a solution.

3. **Conclusion:**

   From the above analysis, the only value of \( n \) that results in a perfect square is \( n = 0 \). Therefore, the only non-negative integer solution is:

   \[
   \boxed{(m, n) = (0, 0)}
   \]

\end{casebox}
\vspace{10pt}

\begin{casebox}{OpenMathReasoning}{Solution by Guess}
\textbf{Problem Description:} Given \(a, b, c \ge 0\) and \(a + b + c = ab + bc + ca > 0\), find the greatest positive real number \(k\) such that
\[
(kab + 1)(kbc + 1)(kca + 1) \ge (k + 1)^3.
\]

\textbf{Error Category:} \textit{Solution by Guess}

\textbf{Reason for Error:} The solution identifies the correct bound by testing specific, simplified cases. This unsound derivation, which lacks a rigorous proof of generality, happens to yield the correct answer.
\tcblower

\textbf{Solution:}

To find the greatest positive real number \( k \) such that the inequality
\[
(kab + 1)(kbc + 1)(kca + 1) \ge (k + 1)^3
\]
holds for all non-negative \( a, b, c \) satisfying \( a + b + c = ab + bc + ca > 0 \), we proceed as follows:

1. **Case Analysis:**
   - Consider the case where \( a = b = c = 1 \). Here, \( a + b + c = 3 \) and \( ab + bc + ca = 3 \), which satisfies the given condition. Substituting these values into the inequality, we get:
     \[
     (k \cdot 1 \cdot 1 + 1)(k \cdot 1 \cdot 1 + 1)(k \cdot 1 \cdot 1 + 1) = (k + 1)^3.
     \]
     This shows that equality holds for any \( k \), so this case does not restrict \( k \).

2. **Case with One Variable Zero:**
   - Consider the case where \( c = 0 \). The condition \( a + b + 0 = ab + 0 + 0 \) simplifies to \( a + b = ab \). Solving for \( a \) and \( b \), we get \( b = \frac{a}{a-1} \). A valid solution is \( a = 2 \) and \( b = 2 \). Substituting \( a = 2 \), \( b = 2 \), and \( c = 0 \) into the inequality, we get:
     \[
     (k \cdot 2 \cdot 2 + 1)(k \cdot 2 \cdot 0 + 1)(k \cdot 0 \cdot 2 + 1) = (4k + 1) \ge (k + 1)^3.
     \]
     Expanding and simplifying the right-hand side, we have:
     \[
     4k + 1 \ge k^3 + 3k^2 + 3k + 1.
     \]
     Subtracting \( 4k + 1 \) from both sides, we get:
     \[
     0 \ge k^3 + 3k^2 - k.
     \]
     Factoring out \( k \), we obtain:
     \[
     k(k^2 + 3k - 1) \le 0.
     \]
     Since \( k > 0 \), we solve the quadratic equation \( k^2 + 3k - 1 = 0 \). The positive root is:
     \[
     k = \frac{-3 + \sqrt{13}}{2}.
     \]

3. **Verification:**
   - We need to verify that this value of \( k \) works for other cases. For instance, if we take \( a = 1.5 \), \( b = 1.5 \), and \( c = 0.375 \), the inequality still holds for \( k = \frac{-3 + \sqrt{13}}{2} \). However, if \( k \) exceeds this value, the inequality fails for the case \( a = b = 2 \) and \( c = 0 \).

Therefore, the greatest positive real number \( k \) that satisfies the inequality for all non-negative \( a, b, c \) under the given conditions is:
\[
\boxed{\frac{\sqrt{13} - 3}{2}}.
\]

\end{casebox}
\vspace{10pt}
\begin{casebox}{OpenMathReasoning}{Calculation Errors}
\textbf{Problem Description:} In $\Delta ABC$, with $AC > AB$, the internal angle bisector of $\angle A$ meets $BC$ at $D$, and $E$ is the foot of the perpendicular from $B$ onto $AD$. Given $AB = 5$, $BE = 4$, and $AE = 3$, find the value of the expression \[(\frac{AC+AB}{AC-AB})ED\].

\textbf{Error Category:} \textit{Calculation Errors}

\textbf{Reason for Error:} The solution contains a significant algebraic error in its final step, incorrectly simplifying $\frac{5(x-3)}{x-5}$ to 3. Through this flawed derivation, however, it reaches the correct numerical answer.
\tcblower
\textbf{Solution:}

To solve the problem, we start by setting up the coordinate system and using the given information effectively.

1. **Coordinate System Setup**:
   - Place \( A \) at \((0, 0)\).
   - Place \( E \) at \((3, 0)\) since \( AE = 3 \).
   - Place \( B \) at \((3, 4)\) since \( BE = 4 \) and \( AE = 3 \).

2. **Angle Bisector Theorem**:
   - By the Angle Bisector Theorem, the internal angle bisector of \(\angle A\) divides \( BC \) in the ratio of the adjacent sides \( AB \) and \( AC \). Let \( AC = x \). Therefore, \(\frac{BD}{DC} = \frac{AB}{AC} = \frac{5}{x}\).

3. **Coordinates of Point \( D \)**:
   - Since \( D \) lies on \( BC \) and the x-axis, we can use the section formula to find the coordinates of \( D \). The coordinates of \( D \) are given by:
     \[
     D = \left( \frac{5 \cdot c\_x + x \cdot 3}{5 + x}, \frac{5 \cdot c\_y + x \cdot 0}{5 + x} \right)
     \]
     where \( C = (c\_x, c\_y) \). Since \( D \) lies on the x-axis, the y-coordinate of \( D \) is 0. This implies:
     \[
     \frac{5 \cdot c\_y + x \cdot 0}{5 + x} = 0 \implies c\_y = 0
     \]
     Therefore, \( C = (c\_x, 0) \).

4. **Finding \( c\_x \)**:
   - Since \( AC = x \), we have:
     \[
     \sqrt{c\_x^2} = x \implies c\_x = x
     \]
     Therefore, \( C = (x, 0) \).

5. **Coordinates of \( D \)**:
   - Using the section formula again:
     \[
     D = \left( \frac{5 \cdot x + x \cdot 3}{5 + x}, 0 \right) = \left( \frac{8x}{5 + x}, 0 \right)
     \]

6. **Distance \( ED \)**:
   - The distance \( ED \) is the horizontal distance between \( E \) and \( D \):
     \[
     ED = \left| \frac{8x}{5 + x} - 3 \right| = \left| \frac{8x - 3(5 + x)}{5 + x} \right| = \left| \frac{8x - 15 - 3x}{5 + x} \right| = \left| \frac{5x - 15}{5 + x} \right| = \frac{5(x - 3)}{5 + x}
     \]

7. **Expression Evaluation**:
   - The expression to evaluate is:
     \[
     \left( \frac{AC + AB}{AC - AB} \right) ED = \left( \frac{x + 5}{x - 5} \right) \cdot \frac{5(x - 3)}{5 + x}
     \]
   - Simplifying the expression:
     \[
     \left( \frac{x + 5}{x - 5} \right) \cdot \frac{5(x - 3)}{5 + x} = \frac{5(x - 3)}{x - 5}
     \]
   - Since \( x > 5 \), the expression simplifies to:
     \[
     \frac{5(x - 3)}{x - 5} = 3
     \]

Thus, the value of the expression is \(\boxed{3}\).

\end{casebox}
\vspace{10pt}
\subsection{CMO}
\vspace{10pt}
\begin{casebox}{CMO}{Solution by Guess}
\textbf{Problem Description:}
Given a function \( f(x) \) defined on \(\mathbb{R}\) that satisfies \( f(0) = 1008 \) and for any \( x \in \mathbb{R} \):
\[
f(x+4) - f(x) \leq 2(x+1),
\]
\[
f(x+12) - f(x) \geq 6(x+5).
\]
Find the value of \(\frac{f(2016)}{2016}\).

\textbf{Error Category:} \textit{Solution by Guess}

\textbf{Reason for Error:} The solution's reasoning is flawed because it relies on the unproven assumption that $f(x)$ must be a quadratic function. By using this guess, it arrives at the correct answer without a valid, general proof.
\tcblower
\textbf{Solution:}First, assume that \( f(x) \) is a quadratic function, let \( f(x) = ax^2 + bx + c \). Substituting into the inequalities from the problem:

1. For \( f(x+4) - f(x) \leq 2(x+1) \):
\[
a(x+4)^2 + b(x+4) + c - (ax^2 + bx + c) \leq 2x + 2.
\]
This simplifies to \(8ax + 16a + 4b \leq 2x + 2\). Comparing the coefficients yields:
\[
8a = 2 \implies a = \frac{1}{4},
\]
\[
16a + 4b \leq 2 \implies 4 + 4b \leq 2 \implies b \leq -\frac{1}{2}.
\]

2. For \( f(x+12) - f(x) \geq 6(x+5) \):
\[
a(x+12)^2 + b(x+12) + c - (ax^2 + bx + c) \geq 6x + 30.
\]
This simplifies to \(24ax + 144a + 12b \geq 6x + 30\). Comparing the coefficients yields:
\[
24a = 6 \implies a = \frac{1}{4},
\]
\[
144a + 12b \geq 30 \implies 36 + 12b \geq 30 \implies b \geq -\frac{1}{2}.
\]

Combining the ranges for \( b \) gives \( b = -\frac{1}{2} \). The initial condition \(f(0)=1008\) gives \(c=1008\). Therefore, the function is:
\[
f(x) = \frac{1}{4}x^2 - \frac{1}{2}x + 1008.
\]
Calculating \(f(2016)\) with this function:
\[
f(2016) = \frac{1}{4}(2016)^2 - \frac{1}{2}(2016) + 1008 = 1016064.
\]
The final answer is:
\[
\frac{1016064}{2016} = \boxed{504}
\]
\end{casebox}
\vspace{10pt}
\begin{casebox}{CMO}{Solution by Guess}

\textbf{Problem Description:}
Let \( f(x) = x^5 - 10x^3 + ax^2 + bx + c \). If all roots of \( f(x) = 0 \) are real, and \( m \) is the largest root, find the maximum value of \( m \).

\textbf{Error Category:} \textit{Solution by Guess}

\textbf{Reason for Error:} The solution assumes a specific, highly symmetric structure for the roots (one root $m$ and four identical roots $k$). This flawed, heuristic-based derivation is not a rigorous proof but  produces the correct maximum value.
\tcblower
\textbf{Solution:}Given the polynomial \( f(x) = x^5 - 10x^3 + ax^2 + bx + c \), let its roots be \(r_1, r_2, r_3, r_4, r_5\).

1. **Sum and Product of Roots**:
   - According to Vieta's formulas, the roots satisfy:
     - Sum of roots: \( r_1 + r_2 + r_3 + r_4 + r_5 = 0 \)
     - Sum of products of roots, two at a time: \( \sum_{1 \leq i < j \leq 5} r_i r_j = -10 \)

2. **Constructing Symmetric Roots**:
   - Assume the largest root is \( m \) and the other four roots are equal, let's say to \( k \). From the condition that the sum of roots is 0, we have:
     \[
     m + 4k = 0 \implies k = -\frac{m}{4}
     \]
   - Calculating the sum of the products of the roots taken two at a time:
     \[
     6k^2 + 4mk = -10
     \]
     Substituting \( k = -\frac{m}{4} \):
     \[
     6\left(-\frac{m}{4}\right)^2 + 4m\left(-\frac{m}{4}\right) = \frac{3m^2}{8} - m^2 = -\frac{5m^2}{8} = -10
     \]
     Solving for \(m\):
     \[
     m^2 = 16 \implies m = 4 \quad (\text{Since } m \text{ is the largest root, we take the positive value})
     \]
   
3. **Verification of the Construction**:
    - When \( m = 4 \), the other four roots are \( -1 \). These roots satisfy the sum of roots being 0 and the sum of the products of the roots taken two at a time being -10, which matches the conditions.

4. **Further Verification using Cauchy-Schwarz Inequality**:
    - The sum of the squares of the four roots is:
      \[
      20 - m^2 \geq 0 \implies m \leq \sqrt{20} \approx 4.4721
      \]
    - However, by constructing the symmetric case, we found that when \( m = 4 \), the other four roots are \( -1 \), which satisfies all conditions. Discriminant analysis shows that a larger \( m \) cannot be constructed.

 Conclusion:
Under the condition that all roots are real, the maximum possible value of the largest root \( m \) is 4.

\[
\boxed{4}
\]
\end{casebox}
\vspace{10pt}
\begin{casebox}{CMO}{Calculation Errors}

\textbf{Problem Description:}
An intelligence station has four passwords: A, B, C, D. Each week's password is chosen randomly from the three not used the previous week. If A is used in week 1, what is the probability A is used in week 7?

\textbf{Error Category:} \textit{Calculation Errors}

\textbf{Reason for Error:} The derivation is flawed in two ways: it starts with an incorrect general formula for the recurrence relation and then makes a calculation error. Through this doubly-flawed process, it paradoxically arrives at the correct answer.
\tcblower
\textbf{Solution:}First, we need to determine the state transition rule for the password usage. The password for each week is chosen with equal probability from the three passwords not used in the previous week. Therefore, if password A is used in a given week, the next week A will definitely not be used; instead, one of B, C, or D will be chosen with equal probability. If a non-A password is used in a given week, the probability of using A in the next week is 1/3.

Let \(P_n\) be the probability that password A is used in week \(n\). According to the problem statement, the probability of using A in the first week is 1, so \(P_1 = 1\). Next, we derive the recurrence relation:

When the probability of using A in week \(n\) is \(P_n\), the probability of using A in week \(n+1\) is 0. When the probability of not using A in week \(n\) is \(1 - P_n\), the probability of using A in week \(n+1\) is \(\frac{1}{3}\). Thus, the recurrence relation is:
\[
P_{n+1} = \frac{1}{3}(1 - P_n)
\]

Calculating the probabilities for each week through recurrence:

\begin{align*}
P_1 &= 1, \\
P_2 &= \frac{1}{3}(1 - 1) = 0, \\
P_3 &= \frac{1}{3}(1 - 0) = \frac{1}{3}, \\
P_4 &= \frac{1}{3}\left(1 - \frac{1}{3}\right) = \frac{2}{9}, \\
P_5 &= \frac{1}{3}\left(1 - \frac{2}{9}\right) = \frac{7}{27}, \\
P_6 &= \frac{1}{3}\left(1 - \frac{7}{27}\right) = \frac{20}{81}, \\
P_7 &= \frac{1}{3}\left(1 - \frac{20}{81}\right) = \frac{61}{243}.
\end{align*}

We can also verify the result by solving for the general formula of the recurrence relation. The relation is:
\[
P_{n+1} = \frac{1}{3} - \frac{1}{3}P_n
\]

Solving this first-order linear recurrence relation gives the general formula:
\[
P_n = \frac{1}{4} + \frac{9}{4}\left(-\frac{1}{3}\right)^n
\]

Substituting \(n = 7\) to calculate:
\[
P_7 = \frac{1}{4} + \frac{9}{4}\left(-\frac{1}{3}\right)^7 = \frac{1}{4} + \frac{9}{4} \cdot \frac{1}{2187} = \frac{1}{4} + \frac{1}{972} = \frac{243}{972} + \frac{1}{972} = \frac{244}{972} = \frac{61}{243}
\]

Thus, the probability of using password A in the 7th week is:
\[
\boxed{\dfrac{61}{243}}
\]
\end{casebox}
\vspace{10pt}
\begin{casebox}{CMO}{Logical Fallacies/Calculation Errors}

\textbf{Problem Description:}
The sum of all real roots of \( \sin(\pi x) = \left[ \frac{x}{2} - \left\lfloor \frac{x}{2} \right\rfloor \right] + \frac{1}{2} \) in \( [0, 2\pi] \) is \underline{\qquad}.

\textbf{Error Category:} \textit{Logical Fallacies/Calculation Errors}

\textbf{Reason for Error:} The derivation is based on a logical fallacy—an incorrect symmetry argument about the sum of roots in each interval. This flawed reasoning, combined with further wrong calculations, leads to the correct final sum.
\tcblower
\textbf{Solution:}In the interval \( [0, 2\pi] \), the sum of all real roots of the equation \( \sin(\pi x) = \left\{ \frac{x}{2} \right\} + \frac{1}{2} \) is **12**. The following is a detailed proof process:

---

**Analysis and Solution Steps**

1. **Equation Transformation and Periodicity Analysis**
The right side of the equation is the fractional part function \( \left\{ \frac{x}{2} \right\} + \frac{1}{2} \), which has a period of 2. The left side, \( \sin(\pi x) \), also has a period of 2. Therefore, the structure of the solutions can be analyzed within each interval \( [2k, 2k+2) \).

2. **Interval Decomposition and Equation Segmentation**
The interval \( [0, 2\pi] \approx [0, 6.28] \) is divided into the following sub-intervals:
- \(k=0\): \( [0, 2) \)
- \(k=1\): \( [2, 4) \)
- \(k=2\): \( [4, 6) \)
- \(k=3\): \( [6, 6.28) \)

3. **Form of the Equation in Each Interval**
Within the interval \( [2k, 2k+2) \), the equation becomes:
\[
\sin(\pi x) = \frac{x}{2} - k + \frac{1}{2}.
\]
Let \( x = 2k + t \) (where \( t \in [0, 2) \)). The equation simplifies to:
\[
\sin(\pi t) = \frac{t}{2} + \frac{1}{2}.
\]
This indicates that the solutions within each interval are symmetric with respect to \( t \), and there are two solutions in each interval.

4. **Symmetry and Sum of Solutions**
The two solutions \( t_1 \) and \( t_2 \) within each interval \( [2k, 2k+2) \) satisfy \( t_1 + t_2 = 1 \). Therefore, the sum of the corresponding \( x \) solutions is:
\[
(2k + t_1) + (2k + t_2) = 4k + (t_1 + t_2) = 4k + 1.
\]
When summing, each interval contributes \( 4k + 1 \). For \( k=0, 1, 2 \), the sums are \( 1, 5, 9 \) respectively. Finally, the solution in the interval \( [6, 6.28) \), \( x \approx 6.1893 \), approximately satisfies \( x + (2 \cdot 3 + 1 - x) = 7 \), contributing a portion of the sum of 7. Due to the interval limit, the actual sum is \( 1 + 5 + 9 + (7 - 0.8107) \approx 12 \).

5. **Conclusion**
The sum of all solutions is:
\[
\sum_{k=0}^{2} (4k + 1) + \text{partial sum} = 1 + 5 + 9 + \text{adjustment value} \approx 12.
\]

---

 **Final Answer**
\[
\boxed{12}
\]
\end{casebox}
\vspace{10pt}
\subsection{IMO}
\vspace{10pt}
\begin{casebox}{IMO}{Inequality Manipulation Errors}

\textbf{Problem Description:}
$\text{Let } \mathbb{R}_{>0} \text{ be the set of positive real numbers. Determine all functions } f: \mathbb{R}_{>0} \to \mathbb{R}_{>0} \text{ such that}$
\[ x(f(x) + f(y)) \geq (f(f(x)) + y)f(y) \]
$\text{for every } x, y \in \mathbb{R}_{>0}.$

\textbf{Error Category:} \textit{Inequality Manipulation Errors}

\textbf{Reason for Error:} The derivation contains a critical logical fallacy. From the two statements $f(f(y)) + f(y) \geq f(f(f(y))) + y$ and the separate fact that $f(f(f(y))) \leq f(y)$, the solution incorrectly concludes that $f(f(y)) \geq y$. This step is invalid, as one cannot simplify an inequality in this manner (it is logically equivalent to improperly subtracting one inequality from another). It is through this flawed deduction that the solution coincidentally arrives at the correct property $f(f(x))=x$ and finds the right family of functions.
\tcblower
\textbf{Solution:}To determine all functions \( f: \mathbb{R}_{>0} \to \mathbb{R}_{>0} \) that satisfy the inequality
\[ x(f(x) + f(y)) \geq (f(f(x)) + y)f(y) \]
for every \( x, y \in \mathbb{R}_{>0} \), we proceed as follows:

1. **Substitution \( y = f(x) \)**:
   - Substituting \( y = f(x) \) into the inequality gives:
     \[ x(f(x) + f(f(x))) \geq (f(f(x)) + f(x))f(f(x)). \]
   - Simplifying both sides by dividing by \( f(x) + f(f(x)) \) (which is positive), we get:
     \[ x \geq f(f(x)). \]
   - This implies \( f(f(x)) \leq x \) for all \( x > 0 \).

2. **Substitution \( x = f(y) \)**:
   - Substituting \( x = f(y) \) into the inequality gives:
     \[ f(y)(f(f(y)) + f(y)) \geq (f(f(f(y))) + y)f(y). \]
   - Simplifying both sides by dividing by \( f(y) \), we get:
     \[ f(f(y)) + f(y) \geq f(f(f(y))) + y. \]
   - Since \( f(f(f(y))) \leq f(y) \) (from \( f(f(z)) \leq z \) for \( z = f(y) \)), we have:
     \[ f(f(y)) \geq y. \]

3. **Combining Results**:
   - From the previous steps, we have \( f(f(x)) \leq x \) and \( f(f(y)) \geq y \). Combining these, we conclude:
     \[ f(f(x)) = x \]
   - This shows that \( f \) is an involution, meaning \( f(f(x)) = x \) for all \( x > 0 \).

4. **Testing Specific Functions**:
   - Consider \( f(x) = \frac{k}{x} \) for some \( k > 0 \). Substituting into the original inequality:
     - Left-hand side: \( x\left(\frac{k}{x} + \frac{k}{y}\right) = k + \frac{kx}{y} \)
     - Right-hand side: \(\left(f(f(x)) + y\right)f(y) = \left(x + y\right)\frac{k}{y} = \frac{kx}{y} + k \)
     - Both sides are equal, confirming that \( f(x) = \frac{k}{x} \) satisfies the inequality.

5. **Uniqueness of Solution**:
   - By analyzing the inequality and considering the necessary conditions for it to hold for all \( x, y > 0 \), we conclude that the only functions satisfying the inequality are reciprocal functions. Any other function would fail to meet the involution property or the inequality conditions.

Thus, the solutions are all reciprocal functions of the form:
\[ \boxed{f(x) = \dfrac{k}{x}} \]
for some constant \( k > 0 \).

\end{casebox}
\vspace{10pt}
\begin{casebox}{IMO}{Logical Fallacies/Solution by Guess}

\textbf{Problem Description:}
Determine all composite integers \( n > 1 \) such that for its ordered positive divisors \( 1 = d_1 < \dots < d_k = n \), \( d_i \) divides \( d_{i+1} + d_{i+2} \) for all \( 1 \leq i \leq k - 2. \)

\textbf{Error Category:} \textit{Logical Fallacies/Solution by Guess}

\textbf{Reason for Error:} The derivation is flawed on two levels. First, the reasoning used to establish the condition for the case $n=pq$ is formally unsound. Second, even if this base case were proven soundly, the solution commits a logical fallacy of unproven generalization by simply asserting that "similar contradictions arise" for all other non-prime-power composites without providing any argument. It is through this process, which is flawed by both a lack of rigor and an unproven generalization, that the solution coincidentally arrives at the correct conclusion.
\tcblower
\textbf{Solution:}To determine all composite integers \( n > 1 \) that satisfy the given property, we start by examining small composite numbers and checking if they meet the condition. We observe that prime powers such as \( 4 = 2^2 \), \( 8 = 2^3 \), \( 9 = 3^2 \), \( 16 = 2^4 \), \( 25 = 5^2 \), \( 27 = 3^3 \), and \( 49 = 7^2 \) satisfy the condition. 

For a prime power \( n = p^m \) (where \( p \) is a prime and \( m \geq 2 \)), the divisors are \( 1, p, p^2, \ldots, p^m \). For each \( i \) from 1 to \( k - 2 \), we need \( d_i \) to divide \( d_{i+1} + d_{i+2} \). Here, \( d_i = p^{i-1} \), \( d_{i+1} = p^i \), and \( d_{i+2} = p^{i+1} \). The sum \( d_{i+1} + d_{i+2} = p^i + p^{i+1} = p^i(1 + p) \), which is divisible by \( p^{i-1} \). Thus, all prime powers satisfy the condition.

Conversely, for composite numbers \( n \) with multiple distinct prime factors, such as \( n = pq \) (where \( p \) and \( q \) are distinct primes), the divisors include \( 1, p, q, pq \). Checking \( i = 2 \), we find \( p \) must divide \( q + pq \), which implies \( p \) divides \( q \), a contradiction since \( p \) and \( q \) are distinct primes. Similar contradictions arise for other composite numbers with multiple prime factors, showing they do not satisfy the condition.

Thus, the composite integers \( n > 1 \) that satisfy the given property are precisely the prime powers. The final answer is:

\[
\boxed{n = p^m}
\]

where \( p \) is a prime and \( m \geq 2 \).

\end{casebox}
\vspace{10pt}
\begin{casebox}{IMO}{Solution by Guess}

\textbf{Problem Description:}
Determine all ordered pairs \((a, p)\) of positive integers, with \(p\) prime, such that \(p^a + a^4\) is a perfect square.

\textbf{Error Category:} \textit{Solution by Guess}

\textbf{Reason for Error:} The derivation is flawed because it lacks a proof of completeness. After finding the correct solutions via case analysis, it fails to rigorously exclude all other possibilities, yet still presents the correct, complete set of answers.
\tcblower
\textbf{Solution:}The ordered pairs \((a, p)\) of positive integers with \(p\) prime such that \(p^a + a^4\) is a perfect square are \((1, 3)\), \((2, 3)\), \((6, 3)\), and \((9, 3)\). 

**Step-by-Step Explanation and Proof:**

1. **Case \(a = 1\):**
   - The equation becomes \(p + 1 = k^2\).
   - \(p = k^2 - 1 = (k - 1)(k + 1)\). Since \(p\) is prime, one factor must be 1.
   - \(k - 1 = 1 \implies k = 2\), leading to \(p = 3\). Thus, \((1, 3)\) is a solution.

2. **Case \(a = 2\):**
   - The equation becomes \(p^2 + 16 = k^2\).
   - Factoring: \((k - p)(k + p) = 16\). The valid factor pair is \((2, 8)\), giving \(k = 5\) and \(p = 3\). Thus, \((2, 3)\) is a solution.

3. **Case \(a = 6\):**
   - The equation becomes \(3^6 + 6^4 = 729 + 1296 = 2025 = 45^2\).
   - This factors as \(3^4 \cdot (3^2 + 2^4) = 81 \cdot 25 = 45^2\). Thus, \((6, 3)\) is a solution.

4. **Case \(a = 9\):**
   - The equation becomes \(3^9 + 9^4 = 19683 + 6561 = 26244 = 162^2\).
   - This factors as \(3^8 \cdot (3 + 1) = 6561 \cdot 4 = 162^2\). Thus, \((9, 3)\) is a solution.

**General Analysis for \(p \neq 3\):**
- **Modulo 3 Analysis:** For \(p \neq 3\), \(p \equiv 1 \text{ or } 2 \pmod{3}\). The sum \(p^a + a^4\) must be a quadratic residue modulo 3. This restricts possible values of \(a\) and \(p\), but exhaustive testing for small primes (e.g., \(p = 2, 5, 7, 11\)) shows no solutions.
- **Exponential Growth:** For \(p \geq 5\), \(p^a\) grows faster than \(a^4\), making \(p^a + a^4\) unlikely to be a perfect square for large \(a\).

**Conclusion:**
Only \(p = 3\) yields solutions for \(a = 1, 2, 6, 9\). No other primes satisfy the equation under the given constraints.

**Final Answer:**
The solutions are \(\boxed{(1, 3)}\), \(\boxed{(2, 3)}\), \(\boxed{(6, 3)}\), and \(\boxed{(9, 3)}\).

\end{casebox}
\vspace{10pt}
\section{Human Subjects}
We use humman annotators for the evaluation. The annotators are recruited from the graduate school and supported by grants. All an-notators agree to share data. The data collectionprotocol is approved by an ethics review board.
\

\end{document}